# Approach for modeling single branches of meadow orchard trees with 3D point clouds

J. Straub[1*], D. Reiser[1] and H. W. Griepentrog[1]

[1]*University of Hohenheim, Institute of Agricultural Engineering, Garbenstr. 9, D-70599, Stuttgart, Germany*
*jonas.straub@uni-hohenheim.de

**Abstract**

The cultivation of orchard meadows provides an ecological benefit for biodiversity, which is significantly higher than in intensively cultivated orchards. The goal of this research is to create a tree model to automatically determine possible pruning points for stand-alone trees within meadows. The algorithm which is presented here is capable of building a skeleton model based on a pre-segmented photogrammetric 3D point cloud. Good results were achieved in assigning the points to their leading branches and building a virtual tree model, reaching an overall accuracy of 95.19 %. This model provided the necessary information about the geometry of the tree for automated pruning.

**Introduction**

Orchard meadows are an essential part of a small-scale cultural landscape, especially in the south of Germany. However, orchard meadows are not currently profitable from an economic point of view and are usually poorly maintained. As a result, more than 80 % of the orchard trees in the state of Baden-Württemberg are currently only irregularly pruned, or are not pruned at all (MLR, 2015). However, for the preservation of the trees, regular pruning is important. Automated pruning could make a meaningful contribution to the preservation of the cultural landscape. In order to be able to implement automated pruning, the branch structure must be detected with the help of sensor technology. In research, there are already a number of projects dealing with the generation of 3D models of orchard trees. For example, Sanz et al. (2018) calculated leaf area based on LiDAR sensor data. Furthermore, a graph-based topologic structure of a tree has already been presented (Arikapudi et al., 2015), which has some parallels to the model used in this paper. Here, a distinction is made between trunk, main branches, and sub-branches in order to build a model as a hierarchical graph based on this classification. A similar classification is used in this paper to distinguish between trunk, major and minor branches. There are already some studies such as Méndez et al. (2016) and Tabb & Medeiros (2017) that use a graph-based representation of an orchard tree for modelling. However, in these studies, as well as in He & Schupp (2018), the focus is on the analysis of orchard trees in plantation cultivations that has different preconditions, which means that the structure of these trees does not reach the same complexity as in an orchard meadow.

The objective of the study reported in this paper was to create a geometrically and topologically accurate skeleton tree model of all relevant branch structures starting from a 3D point cloud. The graph of this 3D model should be hierarchical because branch direction is important for tree pruning. It should be possible to segment and reference the point cloud by using the skeleton tree model.

**Materials and methods**

Overview

Using a photogrammetric tree point cloud, a geometric skeleton of the tree is created, which corresponds to a hierarchical tree structure. The process can be divided into preprocessing and extracting the tree model. The preprocessing is used to reduce the input data to the main tree structure. Afterwards, the tree trunk is segmented based on a Random sample consensus (RANSAC) algorithm (Fischler & Bolles, 1981). The remaining branches are clustered into smaller segments using a k-means algorithm (Elkan, 2003). The resulting segments are used as nodes in a graph forming a virtual tree model. This graph includes the tree structure, but also incorrect connections to neighboring nodes that were then detected and deleted. Therefore, distance parameters are calculated for the edges of the graph as weights. The final graph of the model is determined using the Dijkstra algorithm (Dijkstra, 1959).

Data acquisition and point cloud generation

A single high stem apple tree was recorded as a photogrammetric 3D point cloud. The images were taken with a DSLR APS-C Camera (Nikon D7500 21 MPix) using a 10.5 mm fisheye lens. The image acquisition was made in two circles, at different heights (approx. 1.50 to 1.90 m above ground) around the tree to detect all the relevant structures. The pictures were then processed with the Agisoft Metashape software (Agisoft, 2020) using the automatic calibration of the camera. The structure from motion algorithm generated a dense 3D point cloud of the trees from the images (see Figure 1A). The manually measured circumference of the tree trunk at 0.5 m height was used to scale the point cloud. This scaling was sufficient for the detection of the tree structure, because the relative geometry of the tree is considered. However, for autonomous tree pruning, this scaling factor must be determined more precisely. Most of the preprocessing was done with the Cloud Compare software (CloudCompare, 2020). The model extraction was done in Python using the libraries scikit-learn (Pedregosa et al., 2011) for segmentation, SciPy (Virtanen et al., 2020) for graph computations and NumPy (Harris et al., 2020) for general computation.

Pre-processing

The 3D point cloud was first subsampled to have a consistent point density and to reduce the amount of data. For sub-sampling, a minimum distance of 5 mm between points was chosen. The orientation of the point cloud was adjusted by using a calculated ground plane, based on the RANSAC plane estimation. Next the point cloud was segmented with a Random Forest algorithm (Breiman, 2001) into the classes 'ground' and 'tree', using the distance from the ground plane and the verticality (CloudCompare, 2020). The tree class was then segmented again with the Random Forest classifier, this time to filter out noise caused by the fine structure of the branches, which were photographed against the sky and differ strongly in their color values from the real branch points. The CIELAB color values of each point were used as a feature, as proposed in Riehle et al. (2020). Finally, a SOR (statistical outlier removal) Filter was applied and disconnected

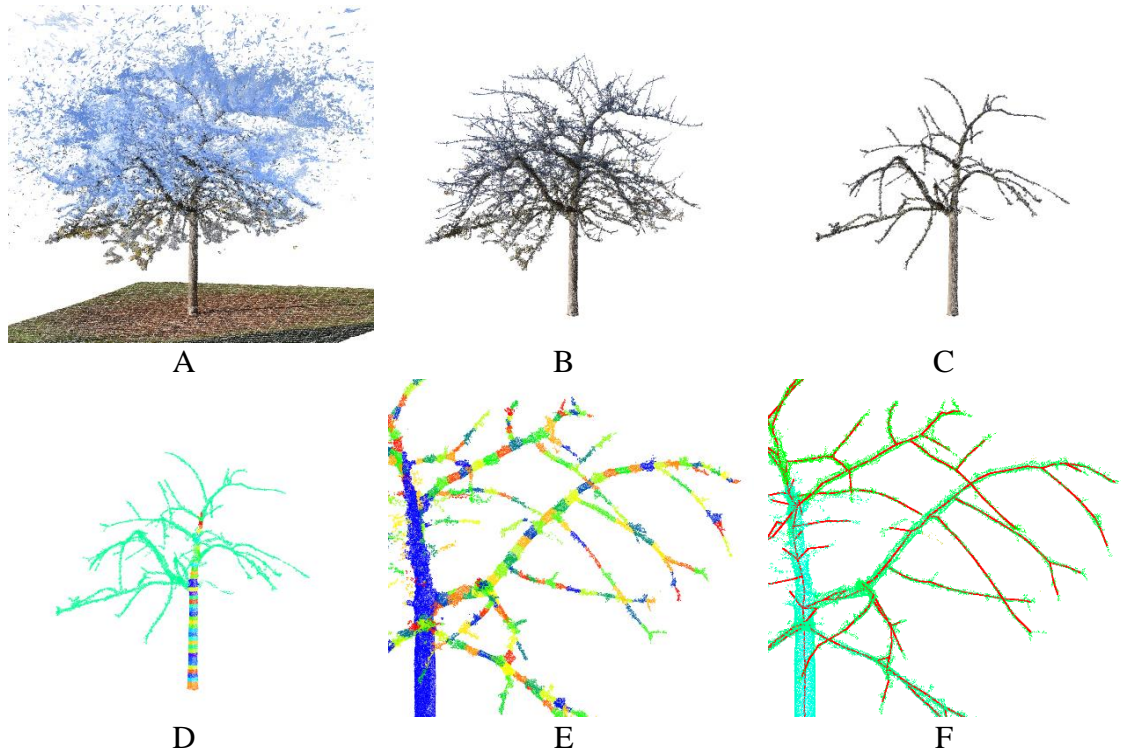

Figure 1. Example of the processing sequence: A – 3D point cloud, B – tree segmented point cloud, C – tree with relevant branches, D – trunk segments in different colours, E – branch segments in different colours, F - tree with extracted tree model

components of the point cloud were removed. The resulting point cloud finally only shows components of the tree branches as seen in Figure 1.B.

As the interest of this research lay in the main branch structure of the upper tree, small branches and tree shoots were disregarded. Therefore, the next step in pre-processing was the segmentation of the tree structure into major and minor branches. A Random Forest algorithm, with 200 trees, a depth of 30 and a minimum sample split of 10, (Breiman, 2001) was applied. This time only geometric features were included (Table 1) as no large derivations in colour between branches of different sizes were detected.

Table 1. List of features used for the random forest classification.

| Feature | Type | Search radius [cm] |
| --- | --- | --- |
| $f_1, f_2, f_3, f_4$ | Number of neighbours | 2, 4, 8, 15 |
| $f_5, f_6, f_7$ | PCA1 | 2, 4, 8 |
| $f_8, f_9$ | PCA2 | 2, 8 |
| $f_{10}, f_{11}, f_{12}, f_{13}$ | Omnivariance | 2, 4, 8, 15 |
| $f_{14}, f_{15}$ | Linearity | 2, 4 |
| $f_{16}$ | Verticality | 15 |
| $f_{17}$ | Planarity | 8 |
| $f_{18}$ | Normal change rate | 8 |
| $f_{19}$ | Surface variation | 8 |
| $f_{20}$ | Cylinder distance | - |

In addition, this made the method independent on the data acquisition method. Beside the cylinder distance feature, all features were available as standard functions calculated in CloudCompare. More information about their calculation can be found in the corresponding documentation (CloudCompare, 2020). These features were selected as the most relevant for the used dataset. For validation of the model, the point cloud was divided into a training (75 %) and a validation dataset (25 %). This classification enabled the separation of the relevant tree structures (Figure 1.C).

Approach

The first step for estimating the tree model was the detection of the tree trunk. This is straight forward because a fitting cylinder for the trunk was calculated previously for $f_{20}$. Points within a distance of 5 cm were assumed to be part of the trunk. These were verified for connectivity to ensure that all points of the trunk were connected to each other. After that, the points were divided vertically into different clusters, with each cluster having an extent in the Z-direction of half the cylinder diameter (as shown in Figure 1.D).

The next step was to cluster the remaining points that should only be associated with significant branches of the tree. A k-means algorithm (Elkan, 2003) was used to divide the point cloud into small branch segments based on the XYZ coordinates. The number of these segments was determined by the voxel volume of the point cloud. The centres of the resulting clusters (Figure 1.E) served as branch nodes for the construction of a graph representing a virtual 3D tree model. The centres of the tree trunk clusters were referred to as trunk nodes. In this case, the number of clusters was defined as 2 clusters per 100 voxels, at a voxel size of 1 cm. All nodes were connected to one graph by building several graphs $A_i$ alongside each other, starting from each of the individual trunk nodes. Distances were calculated between the trunk nodes and the surrounding branch nodes. Then, from each branch node, the distances to the surrounding branch nodes were calculated. From these a combined adjacency matrix with the distances as weights was created to calculate the minimum distance of the respective cluster points. All connections with a distance > 30 mm were removed.

The shortest paths starting from the root trunk node to all branch nodes were calculated using the Dijkstra algorithm (Dijkstra, 1959). Only the shortest paths were used to build the tree structure. The respective trunk node was used as the highest parent node. This initial model was represented by the stack of adjacency matrices $A_i$, which can be combined into a single graph representing the whole tree. This was done by checking all nodes with their input edge, which are present in more than one graph. For these "overlapping" nodes, the distance through the graph to their respective trunk node was calculated. The connection with the smallest graph distance was chosen. The trunk nodes were connected to each other in the Z-direction to complete the graph. The final graph represented the whole tree (Figure 1.F). Each of the nodes of this graph represented a cluster with associated points of the tree point cloud. These points can be assigned to different sub-branches using the graph, which enabled branch-specific segmentation of the point cloud in addition to the skeleton.

Evaluation methods

For the evaluation of the accuracy, a graph was created manually to connect all branch forks of the tree to the trunk. In the computed model, all nodes of the graph that had only two connections were removed. This ensured that there were only branch fork points in the graph. Firstly, a check was made that there was at least one apparently corresponding node of the computed graph in the radius of 50 mm near each node of the reference graph with at least one equivalent connection. However, this did not indicate if the nodes had all the correct connections to each other. For each matched corresponding node in the reference graph, a check was performed to ensure that the connections match with the connections of the computed graph. This had the disadvantage that missing nodes could have a large impact on the result. It was possible that errors appeared disproportionately when looking at the edges, which must be considered in the evaluation. By considering both criteria, a better account was made of the actual correspondence of the computed and reference graphs.

In addition to the evaluation of the graph, the assignment of the points of the point cloud could be used as an indicator of the quality of the result. This was done by checking whether the points assigned to a main branch matched those points in the point cloud previously segmented by hand. This primarily represented a measure of whether points were assigned to the correct main branch, but provides no information about the quality of the finer structures. Therefore, the graph was also verified visually. Although this is not numerically measurable, it allowed a more flexible verification.

**Results and discussion**

Pre-processing
The segmentation between tree and ground reached an overall accuracy (OA) of 97.35 % (Figure 2). Only at the transition between the tree and ground did minor inaccuracies occur, which only had a minimal effect on the further processing. With the subdivision between noise and tree, the focus was on the noise, which is caused by the background sky. The illumination condition played a major role and more data is needed for a good generalization of the model. However, it was a simple and reliable way to remove noise from the point cloud. Crucial for further processing was the segmentation into major and minor branches. Here, an OA of 97.28 % of the validation data was reached. In order to make conclusions about a universal generalization of the classifier, further factors must be considered. A small tree can have a leading branch, which corresponds to the size of only a minor branch of a larger one. Therefore, the model used here had some limitations as to the maximum and minimum tree size. However, since tree size variance was also limited, due to the tree pruning application, the model used is appropriate.

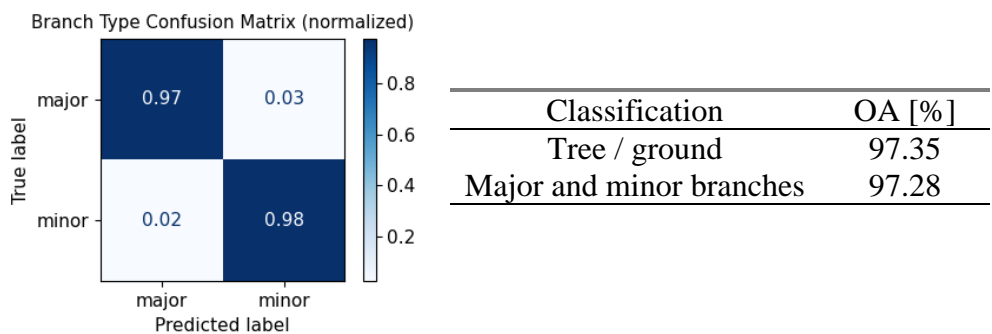

| Classification | OA [%] |
| --- | --- |
| Tree / ground | 97.35 |
| Major and minor branches | 97.28 |

Figure 2. Confusion matrix of the segmentation of the relevant tree structure from the point cloud.

Approach

The results of the node verification achieved a matching accuracy of 81.03 % to the reference graph (Table 2). The results of the edge verification were lower, with a matching of 70.59 % for all edges. If these edges were weighted according to their length, the accuracy was 74.48 %. In the comparison of the individual leading branches, it was noticed that poor correspondence of the nodes was reflected in high errors for the edge detection. This effect was particularly strong at one of the leading branches (LB 3), with a matching accuracy of only 57.14 % and edge detection of 30.77 %. The reason was that missing nodes can have a very strong effect on matching the edges. In this case, this leading branch had only a few nodes, which were closely linked to each other, causing the errors. However, the computed model still provided usable geometry (Figure 3.C). In the further visual verification (Figures 3.A and 3.B), the reconstructed skeleton model matched well with the point cloud and only small errors could be found.

Further results for the model can be seen in Figure 4, where the point assignment to the different leading branches was evaluated. Most of the false assignments were to the trunk class, which was not problematic as the correct connection of the graph was not affected and the transition between trunk and leading branch was not always clearly defined. There were many assignments to the rest class, which meant that parts of a leading branch could not be assigned to any branch. This can also be seen in LB 3 in Figure 3.C and is more problematic as branches can be missing due to a larger hole in the point cloud. With an OA of 95.19 %, the result of the point assignment for the whole tree was very good. Overall, the results showed that this approach was working quite well. However, there is always room for improvement in future work, especially improved robustness against larger holes in the point cloud.

Table 2. Node and edge verification divided by leading branch (LB 1-5), small trunk branches (SB 6) and over all branches.

| | **Node Verification** | | | | | | |
|---|---|---|---|---|---|---|---|
| | LB 1 | LB 2 | LB 3 | LB 4 | LB 5 | SB 6 | all |
| Nodes true [%] | 93.75 | 88.46 | 57.14 | 90 | 94.12 | 69.70 | 81.03 |
| | **Edge Verification** | | | | | | |
| | LB 1 | LB 2 | LB 3 | LB 4 | LB 5 | SB 6 | all |
| Edges true [%] | 93.33 | 76 | 30.77 | 77.78 | 80 | 64 | 70.59 |

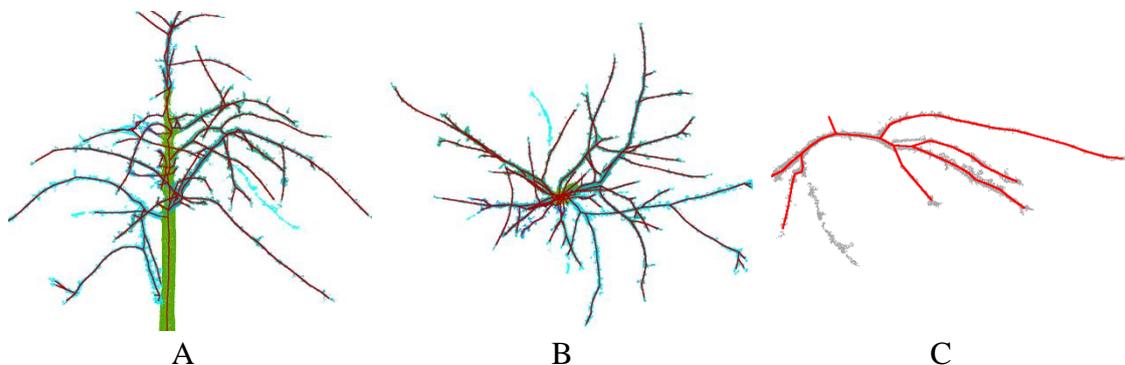

A                                B                                C

Figure 3. A – Tree point cloud segmented by leading branch with skeleton graph B – top view C – leading branch 3 (LB 3) point cloud and calculated skeleton model.

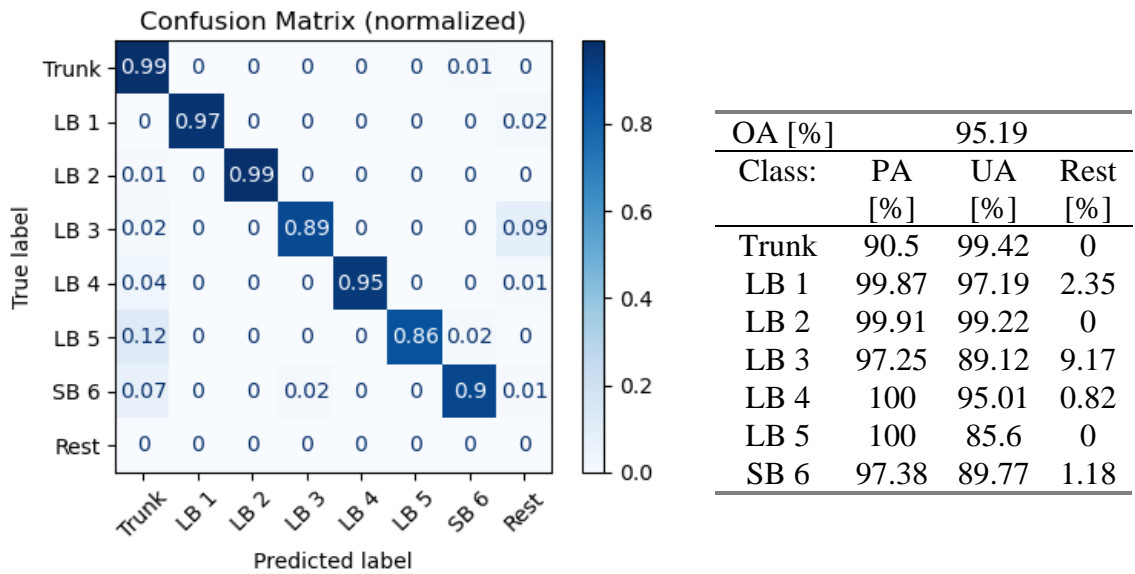

Figure 4. Left - Confusion matrix of the assignment of the points to the trunk, the different leading branches (LB 1-5), the small trunk branches (SB 6) and the unassignable points (Rest). Right - Overall Accuracy (OA), Producers Accuracy (PA), Users Accuracy (UA), Unassignable points (Rest).

Therefore, the unfiltered graph could be used to calculate further features, such as angles between different edges. These angles could then be used for improving the final graph. Furthermore, the graph could be used for the calculation of tree parameters, e.g. branch thickness. In addition, the algorithm, the evaluation and the optimization should be done on a larger number of different trees in the future. Based on a tree model, like the one extracted here, further development for the automation of the pruning of meadow orchard trees can be made. A possibility would be an application in a decision support system, in which such a graph can be used to select pruning points.

**Conclusions**

A photogrammetric point cloud of a meadow orchard tree was acquired. Based on this point cloud, the major tree structures were segmented. From this, a skeleton tree model was extracted as a virtual tree graph. This tree model achieved a node matching of 81.03 %, an edge matching of 70.59 % and an overall accuracy on segmenting the points of the point cloud to the different leading branches of 95.19 %. With the given accuracy, the tree graph was a suitable base for the automation task of tree pruning in an orchard meadow.

**Acknowledgements**

The authors would like to thank the Baden-Württemberg Stiftung for financial support of the research work within the elite program for postdocs.